\documentclass{article}

\usepackage[utf8]{inputenc} 
\usepackage[T1]{fontenc}    
\usepackage{hyperref}       
\usepackage{url}            
\usepackage{booktabs}       
\usepackage{amsfonts}       
\usepackage{nicefrac}       
\usepackage{microtype}      
\usepackage{color}
\usepackage{amsmath}
\usepackage{amsthm}
\usepackage{graphicx}
\usepackage{subcaption}
\usepackage{datetime}

\title{Information-theoretical label embeddings for large-scale image classification}

\author{
  Fran\c{c}ois Chollet \\
  Google, Inc. \\
  \texttt{fchollet@google.com} \\
}

\newdate{date}{01}{06}{2016}
\date{\displaydate{date}}

\begin{document}

\maketitle

\begin{abstract}

We present a method for training multi-label, massively multi-class image classification models, that is faster and more accurate than supervision via a sigmoid cross-entropy loss (logistic regression). Our method consists in embedding high-dimensional sparse labels onto a lower-dimensional dense sphere of unit-normed vectors, and treating the classification problem as a cosine proximity regression problem on this sphere. We test our method on a dataset of 300 million high-resolution images with 17,000 labels, where it yields considerably faster convergence, as well as a 7\% higher mean average precision compared to logistic regression.

\end{abstract}

\section{Introduction}

We consider the problem of predicting to which classes an image belongs, where the number of classes is large (many thousands or tens of thousands) and where each image typically belongs to multiple classes that should all be properly identified: multi-label, massively multi-class classification. In such classification problems, the best practice until now (for instance in use at Google, Inc.) has been to use a deep convolutional neural network such as the ones described in \cite{szegedy15} or \cite{szegedy16}, culminating in a logistic regression layer with a sigmoid cross-entropy loss, with target labels encoded as high-dimensional sparse binary vectors. The use of logistic regression implies an important yet oft overlooked assumption made about the label space: the classes are considered to be statistically independent, each class being treated as an independent dimension in the label space. This is generally not the case in practice: mirroring statistical dependencies found in the real world, label spaces often have a well-defined internal structure, with some labels being more likely to co-occur than other labels. For instance, ``sky'' and ``beach'' are frequently co-occurring labels, while ``crane'' and ``manta ray'' are rarely co-occurring. The sigmoid cross-entropy loss with sparse binary targets does not allow to leverage such observations about the structure of the label space.

There is therefore an opportunity to exploit the internal structure of the label space for gains in training speed, precision, and recall. One simple way to achieve this is to project the labels onto a lower-dimensional manifold --an embedding space-- where a distance function between embedded labels would capture useful statistical dependencies. An appropriate loss function may then allow a parametric model trained via stochastic gradient descent to benefit from the structure of the manifold during training and inference.

\section{Related work}

Several approaches to the problem of learning label embeddings have been proposed in the past. They generally fall into one or more of three broad categories: embeddings capturing label co-occurrences (which is also our approach), visual similarity-based embeddings, and semantic similarity-based embeddings leveraging external sources of data. 

Approaches aimed at label co-occurrence mapping include Compressed Sensing \cite{langford09}, the Error Correcting Output Codes (ECOC) framework \cite{dietterich95}, and well as the randomized linear algebra methods outlined by Mineiro et al. in \cite{mineiro15} and \cite{karampatziakis15}.

Other approaches leverage image similarity together with external sources of semantic knowledge such as word embeddings. These include \cite{zhou15}, \cite{frome13}, \cite{akata15} or \cite{wang16}. \cite{norouzi14} is an approach focused on purely semantic label embeddings.

Additionally, information-theoretical word embedding techniques strongly related to our own approach have long been in use in the natural language processing community. Learning word embeddings by factorizing a pairwise matrix of a correlation-like metric was initially proposed as Latent Semantic Analysis (LSA) in \cite{hofmann99}, then further explored for instance in \cite{bullinaria07}, \cite{baroni10}, and \cite{pennington14}. In 2014, Levy and Goldberg \cite{levy14} showed that the Word2Vec algorithm \cite{mikolov11} was equivalent to factorizing a shifted pointwise mutual information matrix, which is the specific paper from which we drew inspiration for our proposed method.

\section{Embedding label-occurrence via Mutual Information}

Unlike the approaches aiming at capturing either semantic or visual similarity, we are interested purely in the concept of visual co-occurrence of labels, a separate notion. Frequently co-occurring items may be visually dissimilar and may be semantically dissimilar (e.g. clothes and people almost always co-occur in social media images but are not semantically similar). Our approach is therefore most conceptually related to \cite{karampatziakis15} and \cite{mineiro15}, although our method differs significantly both algorithmically and in application scope (as we are interested specifically in image classification, i.e. in visual co-occurrence).

The core idea of our approach is to develop a label embedding space capturing likelihood that any two labels would co-occur in a same picture. Specifically, we focus on the mutual information between these labels as measured from ground-truth annotations.

Let us define basic notations. We consider a set of instances $D$, and a set of classes $C$ of size $N$. To each image $d$ we associate a non-empty subset of $C$, $C^{d}$, the set of classes to which $d$ belongs (``ground-truth annotations''), which we encode as a binary vector $v$ with one or more non-zero values. To each class $C_k (0<=k<N)$ we associate a binary variable $L_{k}$, the presence indicator of class $C_k$ in the ${C^{d}}$ sets. The vectors $v$ are thus the observations of the $L_k$ binary variables over our dataset $D$.

\subsection{Label embedding method}

We consider the matrix of pointwise mutual information $PMI$ \cite{bouma2009normalized} between pairs of $L_k$ variables as measured in the ground-truth observations:

\[ PMI = \log{ \dfrac{ p(L_i, L_j) }{ p(L_i) p(L_j) } } \]

The $PMI$ matrix is symmetric definite positive and can be expressed via singular value decomposition as:

\[ PMI = U \cdot \Sigma \cdot U^{t} \]

where $\Sigma$ is a diagonal matrix, with eigenvalues in the diagonal being ranked from most significant to least significant.

We consider the matrix $E = U \cdot \sqrt{\Sigma}$, so that $PMI = E \cdot E^{t}$. By construction, every row $i$ of $E$ is the projection of a Kronecker $i$ vector (encoding class $i$) onto a latent space where each dimension is an independent factor of variation of the pairwise mutual informations. Because these dimensions are ranked by how much MI variance they explain, the matrix $E$ can be restricted to its $k$ first columns ($k <= n$) at a minimal loss of variance explanation.

The restriction of $E$ to its $k$ first columns, which we note $E^{k}$, is our embedding matrix. Each row ${E^{k}}_i$ is the representation of class $C_i$ in our latent space. Our distance of choice in this space is the cosine distance between different points, which is computationally efficient.

\subsection{Training and inference}

Given an instance $d$ from $D$ and its ground-truth annotation vector $v$ (a sparse binary vector of size $N$, encoding a set of labels), we embed $v$ as $e = E_k \cdot v$, i.e. $e$ the sum of the embeddings of the individual labels in the ground-truth annotations for $d$.

We propose to train a deep convolutional neural network to predict $e$ given $d$. We note the prediction of the network $\tilde{e}$. We use as loss function for the network the cosine proximity between $e$ and $\tilde{e}$ :

\[ loss(e, \tilde{e}) =  -  \dfrac{ e }{ \|e\| } \cdot \dfrac{ \tilde{e}^{t} }{ \|\tilde{e}\| }  \]

At inference time, we decode our predicted embedding vector $\tilde{e}$ into $p$ ranked predictions by sorting all labels by cosine proximity and taking the $p$ closest labels to $\tilde{v}$.

\subsection{Theoretical advantages}

We argue that the proposed method presents several theoretical advantages over logistic regression.

\subsubsection{Zero-shot learning}

If a new class is added to the class set $C$, and if co-occurrence information between this class and other known classes is available, it is possible to give to the new class a position in the embedding space, so that the cosine proximity between the class embedding and the embeddings of known classes approximates the measured mutual information between the classes. Then, by simply taking into account the new class at prediction time (when sorting classes by cosine proximity with the predicted embedding), it is possible to start generating predictions for the new class without having the modify a trained network in any way (zero-shot learning).

\subsubsection{Ground-truth denoising and enrichment}

Our embedding acts as a linking function between labels, hard-coding the statistical dependencies of labels directly into the learning process. If we train a network to associate certain visual feature with the embedding for ``face'', it will automatically learn to associate these features with other labels embedded nearby, such as ``nose'' and ``eyes''. This acts as a label enriching process (adding new ground-truth annotations) and label denoising process (suppressing ground-truth annotations that seem unlikely to co-occur), which is useful in the presence of unreliable annotations.

\subsubsection{Continuity of the target space with regard to the loss function}

In logistic regression, the target space (binary vectors) is not continuous for the loss (sigmoid cross-entropy), i.e. a small change in ground-truth annotations can cause a large change in the corresponding loss value computed by a network. Consider the following example: the picture of a tree might be annotated with ``tree'' and ``branch'', but not with ``trunk'', due to the noise inherent to the ground-truth gathering process. If a model predicts an equal presence probability for all three classes, then adding or removing the ``trunk'' label would cause a 30\% change in the cross-entropy loss value. We argue that one might reasonably expect such lack of continuity to be detrimental to a smooth gradient descent process.

Meanwhile, our proposed method, by embedding labels into a dense continuous space rather than a discrete sparse space, makes the target space continuous for the loss function. Small changes in ground-truth annotations would only result in a correspondingly small movement of the annotations embedding, and thus a small change in cosine proximity loss. Consider our previous example: because the embedding for all three classes ``tree'', ``branch'' and ``trunk'' are very close (due to the extremely frequent co-occurrence of these labels in pictures), removing or adding one of the three labels does not significantly impact the embedding of the final set of of labels, and thus has only a small impact on the cosine proximity loss. See table \ref{closelabels} for relative cosine proximities between these labels.

\section{Experiments}

\subsection{Datasets}

There is no publicly available large-scale image dataset for multi-label, massively multi-class classification. However, Google, Inc. maintains an internal dataset of approximately 300 million web images, nicknamed JFT, first introduced by Hinton in \cite{hinton15}. JFT images are fairly high-resolution (256x256) and are annotated with one or more labels from a set of approximately 17,000 classes. While some of the labels were annotated by hand, most were generated via a variety of heuristics based on semantic context and are therefore somewhat noisy. Associated to this dataset are two smaller, less noisy datasets: Calibration10M, a dataset of 10 million images of which approximately 2.8 million were annotated by hand, and FastEval14k, a dataset of 14,000 picture with very high-quality, dense label annotations (36.5 ground-truth labels per image on average). We use Calibration10M ground-truth annotations for computing label embeddings, we use JFT data for training our cosine proximity regression model, and we evaluate our model on FastEval14k. In table \ref{datasetstats}, we have a look at the label density statistics on Calibration10M and FastEval14k.

\begin{table}[]
\centering
\caption{Classes and labels statistics for Calibration10M and FastEval14k.}
\label{datasetstats}

\resizebox{\columnwidth}{!}{

\begin{tabular}{|l|l|l|l|l|}
\hline
          & \textbf{Classes} & \textbf{Images} & \textbf{Average labels per image} & \textbf{Median occurrences per class}  \\ \hline
\textbf{Calibration10M}  & 17,355   &    2.8M        &                   1.22            &           115                \\ \hline
\textbf{FastEval14k} & 6,112    &     14,000       &                   36.5            &            3                 \\ \hline

\end{tabular}

}
\end{table}

FastEval14k, our evaluation dataset, only contains annotations from approximately 6,000 of the most common labels. Hence, although our embeddings and model are technically trained on all 17,000 labels, the results we show here only include 6,000 labels. Additionally, it is worth noting that the distribution of labels in the dataset on which the embeddings are computed is somewhat different from the distribution of labels in our evaluation dataset. This may mean that our results could be further improved by calibrating the label distribution in the ground-truth used for embedding computation.

\subsection{Metric}

Our metric of choice is Mean Average Precision for top 100 predictions (MAP@100). MAP@100 is defined as:

\[ MAP@100 = \dfrac{1}{N} \sum_{n=1}^{N} \sum_{k=1}^{100} \dfrac{P(k, n)}{100} \]

where $N$ is the number of classes and $P(k, n)$ is the precision for the class $n$ at cut-off $k$ (i.e. precision achieved for the class when only considering the top $k$ predictions made by the model).

Moreover, because we care more about model performance on commonly occurring classes than on rare classes, we associate to each class an importance weight, computed based on the occurrence frequency of the class in a separate dataset of social media images and capped to a minimum and maximum value, and we multiply MAP@100 contributions of the associated labels by this weight. For this reason, the highest achievable MAP@100 in our experiments is higher than 1 (typical values are in the 5-7 range).

\subsection{Process}

\subsubsection{Network architecture}

As our base model, we use a slightly modified version of the Inception v3 architecture described by Szegedy et al. in \cite{szegedy15}, implemented in the TensorFlow framework \cite{tensorflow15} and trained with the RMSprop optimizer \cite{tieleman12}. Our only modifications to the architecture were to remove the dropout layer as well as the auxiliary loss and its associated tower, which act as regularization mechanisms and are not necessary with a very large dataset such as JFT where there is no threat of overfitting in any reasonable amount of time.

Each of our networks was trained on 50 NVIDIA K80 GPUs over several weeks. The longest-running experiment ran for five weeks. All of our models used completely identical configurations for architecture, initialization and optimization --except for the very last layer and its associated loss function.

\subsubsection{Cosine proximity regression}

We first compute an embedding matrix based on ground-truth annotation for Calibration10M and convert ground-truth annotation of JFT to normalized embedded vectors, following the method described earlier. We then train our modified Inception v3 to predict the ground-truth embeddings of the JFT images, using cosine proximity as our loss function as described earlier. At inference time, we use our network to predict embedding vectors for test images, and convert these embeddings into 100 ranked label predictions by taking the taking the 100 closest labels by cosine proximity.

We ran this experiment for three values of $k$, the dimensionality of the embedding space, and eventually settled for $k=4096$.

\subsubsection{Logistic regression}

Our baseline model is the same modified Inception v3 model trained with logistic regression, i.e. ending with 17,000-way sigmoid layer, and supervised by a binary crossentropy loss between the sparse binary ground-truth vectors and the network predictions. From a resources standpoint, the memory footprint and the training-time computational cost of the cosine proximity regression model is noticeably smaller since it features a $k$-dimensional last layer ($k=4,096$) instead of a 17,000-dimensional last layer.

\subsection{Results}

\subsubsection{Qualitative properties of the embedding space}

A qualitative exploration of the embedding space shows that, as expected, classes positioned close together correspond to commonly co-occurring labels (table \ref{closelabels}), which are either synonym labels (e.g. ``flower'' and ``flowering plant'') or labels that are naturally co-occurring (e.g. ``tree'' and ``branches''). The embedding will thus act as a ground-truth completion and enrichment mechanism. More interestingly, our embeddings are capable of performing concept arithmetic in a way somewhat analogous to what Mikolov et al. presented in \cite{mikolov11}. Adding the embeddings of labels A and B will produce a point close to the embeddings of the other labels that also appear when A and B are present, for instance ``man + horse = equestrianism''.

\begin{table}[]
\centering
\caption{Closest labels for some reference labels on the embedding sphere.}
\label{closelabels}

\resizebox{\columnwidth}{!}{%

\begin{tabular}{l l l}

\begin{tabular}{|l|l|}
\hline

\textbf{Closest to: Tree}   & \textbf{Proximity} \\ \hline
Woody plant          & 0.33   \\ \hline
Branch               & 0.23   \\ \hline
Trunk                & 0.22   \\ \hline
Natural environment  & 0.22   \\ \hline
Woodland             & 0.21   \\ \hline
\end{tabular}

\begin{tabular}{|l|l|}              
\hline

\textbf{Closest to: Flower}   & \textbf{Proximity} \\ \hline
Petal             & 0.39  \\ \hline
Flowering plant   & 0.34  \\ \hline
Floristry         & 0.25  \\ \hline
Botany            & 0.24  \\ \hline
Wildflower        & 0.22  \\ \hline
\end{tabular}

\begin{tabular}{|l|l|}
\hline

\textbf{Closest to: Horse}   & \textbf{Proximity} \\ \hline
Horses      & 0.69  \\ \hline
Stallion    & 0.57  \\ \hline
Mustang     & 0.56  \\ \hline
Mare        & 0.54  \\ \hline
Sorrel      & 0.51  \\ \hline
\end{tabular}

\end{tabular}

}
\end{table}

\subsubsection{Effect on classification performance}

\begin{figure}[!ht]
  \caption{Test class-weighted MAP@100 for the embeddings regression approach and the baseline (logistic regression).}
  \label{mapgraphs}
  \centering
    \includegraphics[width=0.85\textwidth]{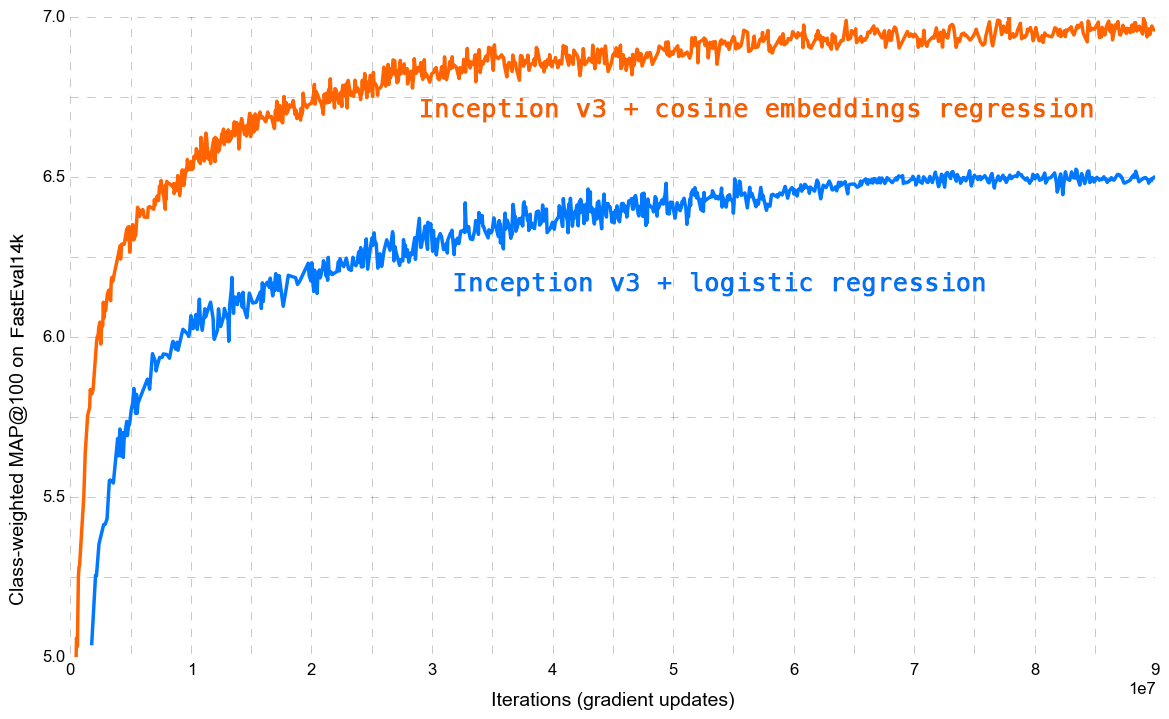}
\end{figure}

Our cosine proximity regression model converges towards a 7\% higher class-weighted MAP@100 (6.95 vs. 6.50), and does so considerably faster (fig. \ref{mapgraphs}). It reaches the final performance of the logistic regression model in only 9 million iterations, i.e. in approximately 10x less iterations than the baseline.

We believe that the faster convergence can be explained by the ground-truth enrichment mechanism implemented by our embedding space, as well as the continuity of the target space with regard to our loss (leading to smoother and thus faster gradient descent). However such claims cannot be rigorously proven at this point.

\section{Conclusions}

We presented a training method for leveraging label-space structure for multi-label, massively multi-class image classification problems. The method has several theoretical advantages over logistic regression, and in practice it proves to perform significantly better than logistic regression on the JFT dataset, leading to significantly faster convergence and 7\% higher mean average precision. These results suggest that our method is a strong alternative to logistic regression in cases where the label space is large and highly structured.

\begin{small}

\bibliographystyle{abbrv}
\bibliography{biblio}

\end{small}

\end{document}